\documentclass[conference]{IEEEtran}
\IEEEoverridecommandlockouts
\usepackage{amsmath,amssymb,amsfonts}
\usepackage{algorithmic}
\usepackage{graphicx}
\usepackage{textcomp}
\usepackage{xcolor}
\def\BibTeX{{\rm B\kern-.05em{\sc i\kern-.025em b}\kern-.08em
    T\kern-.1667em\lower.7ex\hbox{E}\kern-.125emX}}

\usepackage{multirow}

\usepackage{soul}
\usepackage[outline]{contour}
\usepackage{booktabs}
\usepackage{tikz}
\usetikzlibrary{arrows.meta, arrows, patterns}
\usetikzlibrary{calc}
\usepackage{pgfplots}
\pgfplotsset{compat=newest}

\definecolor{tab:darkviolet}{RGB}{148, 0, 211}
\definecolor{tab:bggrey}{RGB}{163, 163, 163}
\definecolor{tab:k}{RGB}{0, 0, 0}
\definecolor{tab:hotpink}{RGB}{255, 105, 180}
\definecolor{tab:black}{RGB}{148, 0, 211}
\definecolor{tab:r}{RGB}{255, 0, 0}
\definecolor{tab:b}{RGB}{1, 1, 255}
\definecolor{tab:olivedrab}{RGB}{107, 142, 35}
\definecolor{tab:deepskyblue}{RGB}{0, 191, 255}
\definecolor{tab:lime}{RGB}{0, 255, 0}
\definecolor{tab:magenta}{RGB}{255,1, 254}
\definecolor{tab:limegreen}{RGB}{50, 204, 51}
\newcommand \colorindicator[1]{%
	\begingroup%
	\setul{0.25ex}{0.4ex}%
	\contourlength{0.2ex}%
	{\textcolor{#1}{\tiny{$\blacksquare \hspace{-.5mm} \blacksquare$}}}%
	\endgroup
}

\makeatletter
\def\ps@IEEEtitlepagestyle{%
  \def\@oddhead{\mycopyrightnotice}%
  \def\@oddfoot{}
  \def\@evenhead{\@IEEEheaderstyle\thepage\hfil\leftmark\hbox{}}\relax
  \def\@evenfoot{}%
}
\def\mycopyrightnotice{%
  \begin{minipage}{\textwidth}
  \centering \scriptsize
  \textcolor{red}{Copyright~\copyright~2023 IEEE. Personal use of this material is permitted. Permission from IEEE must be obtained for all other uses, in any current or future media, including reprinting/republishing this material for advertising or promotional purposes, creating new collective works, for resale or redistribution to servers or lists, or reuse of any copyrighted component of this work in other works. Published in: Proceedings of the 2023 Latin American Robotics Symposium (LARS), 2023 Brazilian Symposium on Robotics (SBR), and 2023 Workshop on Robotics in Education (WRE).}
  \end{minipage}
}
\makeatother

\begin{document}

\title{Motion Consistency Loss for Monocular Visual Odometry with Attention-Based Deep Learning\\

\thanks{
André Françani was supported by CAPES -- Coordination of Improvement of Higher Education Personnel under grant 88887.687888/2022-00. Marcos Maximo is partially funded by CNPq –- National Research Council of Brazil through grant 307525/2022-8.
}
}

\author{\IEEEauthorblockN{1\textsuperscript{st} André O. Françani}
\IEEEauthorblockA{\textit{Autonomous Computational Systems Lab (LAB-SCA)}\\
\textit{Computer Science Division} \\
\textit{Aeronautics Institute of Technology}\\
São José dos Campos, SP, Brazil \\
andre.francani@ga.ita.br}
\and
\IEEEauthorblockN{2\textsuperscript{nd} Marcos R. O. A. Maximo}
\IEEEauthorblockA{\textit{Autonomous Computational Systems Lab (LAB-SCA)} \\ 
\textit{Computer Science Division} \\
\textit{Aeronautics Institute of Technology}\\
São José dos Campos, SP, Brazil \\
mmaximo@ita.br}
}


\maketitle

\IEEEpubidadjcol
\let\thefootnote\relax\footnote{\\979-8-3503-1538-7/23/\$31.00\textcopyright2023 
IEEE}

\begin{abstract}
Deep learning algorithms have driven expressive progress in many complex tasks. The loss function is a core component of deep learning techniques, guiding the learning process of neural networks.
This paper contributes by introducing a consistency loss for visual odometry with deep learning-based approaches. The motion consistency loss explores repeated motions that appear in consecutive overlapped video clips. Experimental results show that our approach increased the performance of a model on the KITTI odometry benchmark.
\end{abstract}


\section{Introduction}
\label{sec:intro}
Deep learning (DL) techniques have shown to be state-of-the-art in diverse complex applications, such as computer vision (CV) and natural language processing (NLP). As evidence, DL models can be employed in image classification tasks \cite{dosovitskiy2020image}, video surveillance \cite{dos2021performance}, text translation \cite{vaswani2017attention}, and others \cite{prangemeier2022yeast}. 

A fundamental component of deep learning techniques is the loss function, being a core element in the optimization algorithm. The loss function is minimized during the training step and the model's parameters are adjusted according to the error between the model's prediction and the expected ground truth. Different loss functions guide the training to a specific task, tuning the model's parameters to a good performance in regression or classification tasks. This means that the choice of the loss function impacts what the model is in fact learning. The focal loss \cite{lin2017focal} is a clear example where the loss function played an important role in object detection applications, enabling inferences as fast as one-stage detectors while reaching accuracy competitive to two-stage detectors. 

A consistency loss may introduce a gain of information capable of making the model more robust and more reliable, increasing the performance. Thus, a consistency loss may guide the learning process to be more accurate in its predictions regarding a particular task, reducing the error between the predicted output and the expected ground truth. Zhu \emph{et al.} \cite{zhu2017unpaired} achieved outstanding qualitative results in generative models by introducing a cycle consistency loss in the image-to-image translation task. 

In this paper, our contribution is a novel consistency loss function that deals with repeated motion in overlapped clips in the context of monocular visual odometry with deep learning. We evaluate our method on the KITTI odometry benchmark \cite{Geiger2012CVPR}, showing that our approach increases the performance of the model.  

The remaining of this paper is organized as follows. Section~\ref{sec:background} provides theoretical background about visual odometry and attention mechanisms. Section~\ref{sec:method} introduces the proposed motion consistency loss. Section~\ref{sec:evaluation} describes the experimental setups. Section~\ref{sec:results} shows the experimental results on the KITTI odometry benchmark. Finally, Section~\ref{sec:conclusion} concludes and shares our ideas for future work.

\section{Background}
\label{sec:background}

\subsection{Monocular visual odometry}

Visual odometry (VO) is widely applied in robotics and autonomous vehicles to estimate the camera's pose given a sequence of image frames \cite{scaramuzza2011VO}. The motion between consecutive time steps $k-1$ and $k$ comprises a rotation matrix $\mathbf{R}_{k} \in S \! O(3)$ and a translation vector $\mathbf{t}_k \in \mathbb{R}^{3 \times 1}$. Note that $\mathbf{R}_{k}$ and $\mathbf{t}_k$, respectively, depict the rotation and translation from time step $k-1$ to $k$. The complete motion can be written as a transformation $\mathbf{T}_k \in \mathbb{R}^{4\times4}$, defined as  

$$\mathbf{T}_k = \begin{bmatrix}
                \mathbf{R}_k & \mathbf{t}_k  \\
                \mathbf{0}   & \mathbf{1}  \\
                \end{bmatrix}. $$
                
\begin{figure}[!tbh]
\centering
\includegraphics[width=0.21\textwidth]{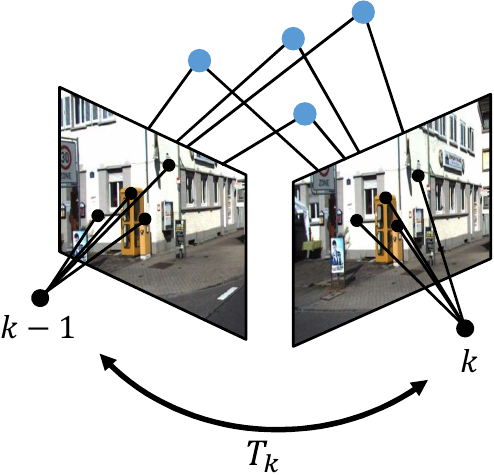}
\caption{Camera's motion between consecutive time steps.}
\label{fig:motion_estimation}
\end{figure}

Furthermore, the poses in VO problems are typically described through 6 degrees of freedom (6-DoF): three rotational, and three translational.
To address the visual odometry problem of estimating the 6-DoF, one can rely on traditional geometry-based approaches \cite{Geiger2011IV, orbslam2}, deep learning-based approaches through end-to-end architectures \cite{wang2017deepvo, li2018undeepvo}, and hybrid methods that mix deep learning-based and geometry-based algorithms depending on the component of the visual odometry \cite{zhan2020visual, francani2022dense}. 
Fig.~\ref{fig:motion_estimation} illustrates the camera's motion between consecutive time steps. In Fig.~\ref{fig:motion_estimation}, the blue dots are keypoints in the scenario, which are used as a reference to estimate the motion via a geometry-based method. 

In the monocular visual odometry case, where the image frames are captured from a single camera, the motion can be obtained up to a scale, resulting in scale ambiguity. This is caused by the loss of depth information when projecting three-dimensional (3D) objects onto the two-dimensional (2D) image space. The accumulation of scale errors over time is called scale drift, being a crucial factor in decreasing the accuracy in monocular visual odometry systems \cite{scaramuzza2011VO}.

\subsection{Time-space attention in monocular visual odometry}

Deep learning-based methods are able to estimate the 6-DoF camera's pose directly from RGB images. In an end-to-end manner, the models estimate the pose given only the sequence of frames as input. DeepVO \cite{wang2017deepvo} is an example of a deep learning-based method that uses a convolutional neural network (CNN) to extract features from consecutive image frames, as well as long short-term memory cells (LSTM) to deal with the temporal information. Furthermore, unsupervised learning techniques can also be employed, as in the UnDeepVO \cite{li2018undeepvo} that uses depth estimation to recover the scale.

For many years, LSTM cells have dominated tasks involving temporal information. However, architectures based on attention mechanisms, such as the Transformer architecture \cite{vaswani2017attention}, have outperformed the accuracy of LSTM models, and have become state-of-the-art in NLP applications.  


The input of the attention block is composed of the query, key, and value vectors, all with dimension $d$. Then, those vectors from different inputs are packed together into the matrices $\mathbf{Q}$, $\mathbf{K}$, and $\mathbf{V}$, which are the inputs of the attention layer. The processing steps can be defined as the Attention function \cite{vaswani2017attention}:

\begin{equation}
\text{Attention}(\mathbf{Q},\mathbf{K},\mathbf{V}) = \text{Softmax}\left(\frac{\mathbf{Q}\mathbf{K}^T}{\sqrt{d}}\right)\mathbf{V},    
\end{equation}
where $\text{Softmax}$ is the well-known $\text{Softmax}$ function. 

Another important structure is the multi-head self-attention layer (MHSA), which has $N_h$ concatenated attention layers running in parallel \cite{vaswani2017attention}.
Since then, the deep learning community has started building models based on the decoder and encoder blocks of the Transformer, and new state-of-the-art networks have begun to emerge, such as BERT \cite{devlin2018bert} and GPT \cite{radford2018improving}. The Transformer architecture also achieved state-of-the-art performance in image recognition benchmarks \cite{dosovitskiy2020image} and video understanding tasks \cite{Arnab_2021_ICCV, gberta_2021_ICML}. For vision tasks, the images are decomposed into $N$ non-overlapping patches of size $P \times P$. Each patch is flattened and embedded into tokens that are sent as input to the network \cite{dosovitskiy2020image}.

Video understanding tasks require space and time information to recognize actions or events in videos. To address this problem with the attention mechanism, G. Bertasius \emph{et al.} \cite{gberta_2021_ICML} proposed mainly two different self-attention architectures to extract spatio-temporal features. One of the proposed architectures is the ``joint space-time'' self-attention, which relates all tokens in space and time together. The other one, more efficient computationally in terms of processing complexity, is the ``divided space-time'' self-attention. It applies first the attention over tokens with the same spatial index (temporal axis) followed by the attention over the tokens from the same frame (spatial axis). 

In the context of visual odometry, our previous work \cite{francani2023tsformer} uses a Transformer-based network with the ``divided space-time'' to estimate the 6-DoF camera's pose for the monocular case. The TSformer-VO is based on the TimeSformer architecture with ``divided space-time'' self-attention in its encoding blocks. The model receives clips with $N_f$ frames of size $C \times H \times W$ as input, where $C$, $H$, and $W$ are the number of channels, the height, and the width of the image frames, respectively. Each clip is divided into $N = HW/P^2$ patches, and each patch is embedded into tokens through a linear map. Finally, the outputs of the embedding layer are the input tokens denoted as $\mathbf{z}^{l}_{(s, t)} \in \mathbb{R}^{E_d}$, where $E_d$ is the embedding dimension of the flattened patch at spatial location $s$, time index $t$, and encoding block $l$. Notice that the model has $L_x$ encoder blocks stacked, indicating the depth of the network in terms of encoding layers.

\begin{figure}[!tbh]
\centering
\includegraphics[width=0.32\textwidth]{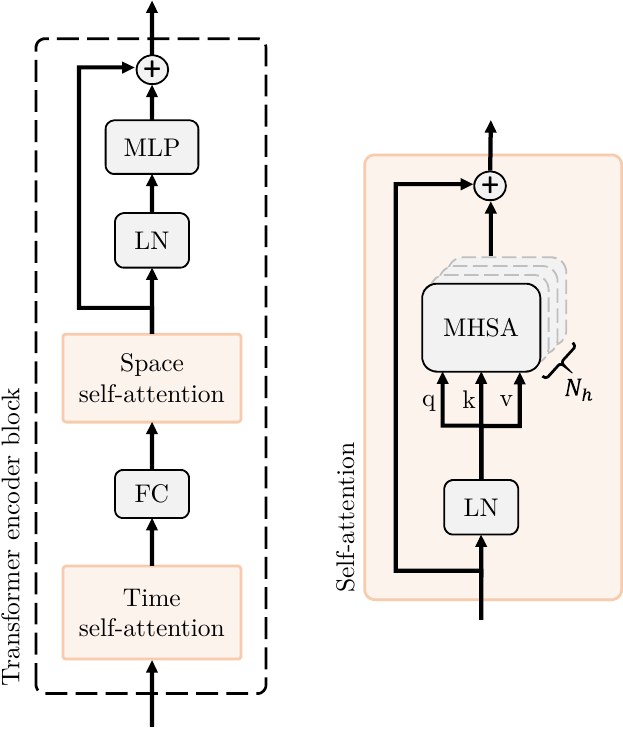}
\caption{Encoder block with ``divided space-time'' self-attention.}
\label{fig:space-time-attn}
\end{figure}

The encoder blocks follow the traditional layers present in the Transformers encoders, namely the layer normalization (LN), multi-head self-attention (MHSA), residual connections, fully-connected layer (FC), and multilayer perceptron (MLP). For the specific ``divided space-time'' self-attention, those layers are related as follows: 
\begin{equation}
    \begin{split}
        \mathbf{a}^{l}_{t} & =  \text{MHSA}\left(\text{LN}\left(\mathbf{z}_{(s,t)}^{l-1}\right)\right) + \mathbf{z}_{(s,t)}^{l-1}, \\
        \mathbf{a}^{l}_{t_{FC}} & = \text{FC}\left( \mathbf{a}^{l}_{t} \right), \\
        \mathbf{a}^{l}_{s} & = \text{MHSA}\left(\text{LN}\left(\mathbf{a}^{l}_{t_{FC}}\right)\right) + \mathbf{a}^{l}_{t_{FC}}, \\
        \mathbf{z}_{(s,t)}^{l} & = \text{MLP}\left(\text{LN}\left(\mathbf{a}^{l}_{s}\right)\right) + \mathbf{a}^{l}_{s}. \\
    \end{split}
\label{eq:encoder}
\end{equation}
Further details about the exact computation of the query, key, and value can be found in \cite{vaswani2017attention, gberta_2021_ICML}. Fig.~\ref{fig:space-time-attn} contains the block diagram of the encoder block with ``divided space-time'' self-attention, summarizing \eqref{eq:encoder}.

\section{Proposed Method}
\label{sec:method}

\begin{figure*}[!tbh]
\centering
\includegraphics[width=0.7\textwidth]{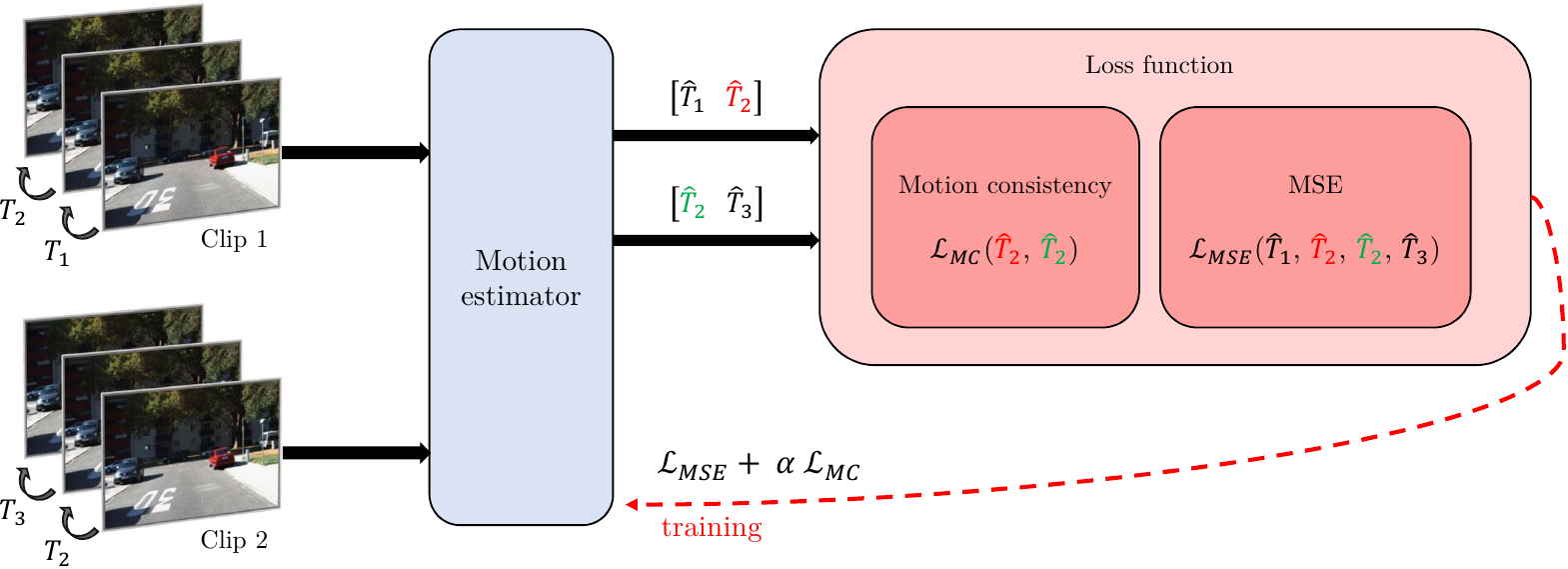}
\caption{Schematic representation of the proposed method. Input clips with overlapped frames go to a model that estimates the camera's motion between consecutive time steps. During the training step, the motion consistency loss is calculated using the estimated motions that appear in all input clips ($\hat{\mathbf{T}}_2$ in this example). The motion consistency loss is weighted by a hyperparameter $\alpha$, and this result is added to the MSE of all estimations. The final loss propagates back to the model during the training step.}
\label{fig:prop_method}
\end{figure*}

The proposed method relies on providing extra information during the training step to increase overall performance. The information gain comes from the fact that we are sampling consecutive clips with $N_f -1$ overlapped frames. This means that the same motion is estimated from different input clips, as shown in Fig.~\ref{fig:prop_method}. With this in mind, we can conduct the learning process of the network by introducing a consistency loss. The idea behind our approach is that consecutive clips should estimate the same motion where they overlap, that is, consecutive clips contain common motions due to the overlapping introduced by sampling the clips. Without the consistency loss, it is not directly clear to the network that consecutive clips contain similar information, once our approach uses the information of the entire clip to infer all the poses in a pair of consecutive frames.

The main part of the total loss function is the mean squared error (MSE) between all predicted motions and their ground truths, given by:
\begin{equation}
    \mathcal{L}_{\textsc{}{MSE}} = \frac{1}{N_f -1} \sum_{w=1}^{N_f-1} {\Vert \mathbf{y}_{w}^{k} - \hat{\mathbf{y}}_{w}^{k} \Vert_{2}^{2}},
\end{equation}
where $\Vert \cdot \Vert_{2}^{2}$ is the squared L2 norm, $\mathbf{y}_{w}^{k}$ is the flattened 6-DoF relative pose at position $w$ of the clip at instant $k$, and $\hat{\mathbf{y}}_{w}^{k}$ is its estimate by the network. When dealing with batch processing, the loss is reduced by the mean over all $B_s$ batch elements.

The second part of the total loss function is our motion consistency loss $\mathcal{L}_{\textsc{}{MC}}$. It is the sum of the squared errors of all predicted motions that appear in more than one input clip, that is, the same motion but from different clips. Therefore, the $\mathcal{L}_{\textsc{}{MC}}$ is defined as:
\begin{equation}
    \mathcal{L}_{\textsc{}{MC}} = \sum_{j=1}^{2N_f-5}\sum_{n=m+1}^{\gamma}\sum_{m=\mu}^{\lambda}
    {\Vert \hat{\mathbf{y}}_{k-j}^{k-m} - \hat{\mathbf{y}}_{k-j}^{k-n} \Vert_{2}^{2}},
\end{equation}
where $\mu = \max{(j-N_f+2, 0)}, \lambda = \min{(N_f-3,j-1)}$, and $\gamma = \min{(N_f - 2, j)}$.

The final loss is the sum of the MSE of all estimations and the motion consistency loss, this one weighted by a hyperparameter $\alpha$ to measure and balance the influence of the consistency loss in the training process. Therefore, the final loss is defined as follows: 
\begin{equation}
    \mathcal{L} = \mathcal{L}_{\textsc{}{MSE}} + \alpha \mathcal{L}_{\textsc{}{MC}}.
\end{equation}

\section{Experimental Setup}
\label{sec:evaluation}

This section contains further information on the model configurations we tested in our experiments, the training setup, the dataset we employed, as well as the evaluation metrics we chose to compare the performance of the models.

\subsection{Model setup}

This work aims to study the influence of the motion consistency loss on the model's performance. Therefore, different architectures were not evaluated, and the model hyperparameters were selected based on our previous work \cite{francani2023tsformer}. All three models in Table~\ref{tab:models} have $L_x = 12$ encoders, each one with $N_h = 6$ heads in the MHSA layer. The patch size is $P=16$, and the embedding dimension is set to $E_d = 384$. Moreover, we tested our approach for clips with $N_f = 3$ frames, meaning that consecutive clips have two overlapped frames, i.e. one motion in common. 

We tested our approach considering different values for $\alpha$ to weight the motion consistency loss. Table~\ref{tab:models} shows the definition of the models according to their loss functions.
\begin{table}[!b]
\caption{Definition of the models with their corresponding loss function}
\label{tab:models}
\centering
\begin{tabular}{cl}
\hline
\textbf{Model} & \textbf{Loss function} \\ \hline
A & $\mathcal{L}_{\text{MSE}}$ \\
B & $\mathcal{L}_{\text{MSE}} + \mathcal{L}_{\text{MC}} $ \\
C & $\mathcal{L}_{\text{MSE}} + 10 \mathcal{L}_{\text{MC}} $ \\ \hline
\end{tabular}
\end{table}

\subsection{Training setup}

We used a computer with a Intel i9-7900X CPU 3.3GHz CPU and a GeForce GTX 1080 Ti GPU with 11GB VRAM for computations, and all the deep learning procedures were implemented with PyTorch 1.10. The models were trained from scratch using random initialization for 95 epochs with Adam optimization. The learning rate was set to $1\times 10^{-5}$, and all the other hyperparameters of the Adam algorithm were the default values. 

Aiming to recycle the code from our previous work \cite{francani2023tsformer}, we sampled the clips with a sliding window of size $2 N_f$ and stride $1$. After that, we shuffled the sampled clips and created batches of size $B_s = 2$. This way, clips in the same batch are not consecutive to each other. However, our approach requires consecutive clips only for the training step to calculate the motion consistency loss. To solve this issue, we split the batches in half, each one containing clips with $N_f$ frames. The main advantage of this sampling process is that we ensure that the clips in the same batch are still shuffled while having their consecutive clips in a second batch. After that, we concatenated both small batches into only one to take advantage of the batch processing ability of deep learning frameworks. Furthermore, this helped us to fit our GPU memory capacity.

\subsection{Dataset}

We developed our algorithms considering the KITTI dataset \cite{Geiger2012CVPR}, which is a benchmark for evaluating visual odometry algorithms. The imagery is recorded from a stereo camera on a moving vehicle. The scenario varies from roads and streets on a normal day in the city, and the car's speed varies from 0 to 90 km/h, challenging visual odometry algorithms at high-speed scenarios. To make the system monocular for our study case, we selected only the RGB images recorded by the left camera. In the KITTI dataset, there are 22 different sequences with different lengths corresponding to different rides. However, the camera's position and orientation over time are only available for 11 of them.

All image frames were resized to $192 \times 640$, keeping the aspect ratio of the dataset while making the dimensions multiple of the patch size ($P=16$) to divide the frames into patches. Furthermore, sequences 00, 02, 08, and 09 are used as training data since they are the four largest recordings. Consequently, the test data is composed of sequences 01, 03, 04, 05, 06, 07, and 10, following the choice in \cite{wang2017deepvo}.

\subsection{Evaluation metrics}

We use the Python KITTI evaluation toolbox\footnote{https://github.com/Huangying-Zhan/kitti-odom-eval} to compare the performance of the models. The metrics are defined as follows:
\begin{itemize}
    \item $t_{err}$: average translational error, given in percentage (\%);
    \item $r_{err}$: average rotational error, given in degrees per 100 meters (º/100 m);
    \item ATE: absolute trajectory error, measured in meters;
    \item RPE: relative pose error for rotation and translation, measured frame-to-frame in degrees (º) and in meters, respectively for the rotation and the translation.
\end{itemize}
According to \cite{Geiger2012CVPR}, the average rotational and translational errors are computed for all subsequences of length $(100, 200, \dots, 800)$ meters.

Prior works evaluate monocular methods with an optimization transformation to align the predictions with the ground truths \cite{orbslam2, zhan2020visual}. Thus, our results are under the 7-DoF alignment due to the scale ambiguity in monocular systems.

\section{Experimental results}
\label{sec:results}

We evaluated the performance of the models in Table~\ref{tab:models} using the KITTI dataset. After the training step, we obtained the quantitative results present in Table~\ref{tab:results}. For each sequence, the best values of each evaluation metric are highlighted in bold, and the second-best values are underlined.
\begin{table*}[!t]
\caption{Quantitative results for the 11 KITTI sequences with ground truth.}
\label{tab:results}
\centering
\resizebox{0.97\textwidth}{!}{
\begin{tabular}{cllllllllllll}
\hline
 & \multicolumn{1}{c}{} & \multicolumn{11}{c}{\textbf{Sequence}} \\ \cline{3-13} 
\multirow{-2}{*}{\textbf{}} & \multicolumn{1}{c}{\multirow{-2}{*}{\textbf{Model}}} & \multicolumn{1}{c}{{\color[HTML]{000000} \textbf{00}}} & \multicolumn{1}{c}{{\color[HTML]{000000} \textbf{01}}} & \multicolumn{1}{c}{{\color[HTML]{000000} \textbf{02}}} & \multicolumn{1}{c}{{\color[HTML]{000000} \textbf{03}}} & \multicolumn{1}{c}{{\color[HTML]{000000} \textbf{04}}} & \multicolumn{1}{c}{{\color[HTML]{000000} \textbf{05}}} & \multicolumn{1}{c}{{\color[HTML]{000000} \textbf{06}}} & \multicolumn{1}{c}{{\color[HTML]{000000} \textbf{07}}} & \multicolumn{1}{c}{{\color[HTML]{000000} \textbf{08}}} & \multicolumn{1}{c}{{\color[HTML]{000000} \textbf{09}}} & \multicolumn{1}{c}{{\color[HTML]{000000} \textbf{10}}} \\ \hline
 & A & 9.053 & {\ul{31.116}} & \textbf{3.105} & \textbf{9.867} & 8.029 & {\ul{12.800}} & 27.212 & {\ul{23.475}} & 5.498 & {\ul{5.063}} & 18.544 \\
 & B & \textbf{3.609} & 41.534 & {\ul{4.084}} & {\ul{10.869}} & {\ul{6.218}} & \textbf{12.263} & {\ul{25.554}} & 25.224 & \textbf{3.283} & \textbf{4.099} & \textbf{12.188} \\
\multirow{-3}{*}{\begin{tabular}[c]{@{}c@{}}$t_{err}$\\ (\%)\end{tabular}} & C & {\ul{4.942}} & \textbf{28.324} & 5.062 & 15.651 & \textbf{4.316} & 13.198 & \textbf{22.334} & \textbf{18.300} & {\ul{5.028}} & 5.785 & {\ul{12.602}} \vspace{0.6mm} \\ \hline
 & A & {\ul{2.018}} & 6.885 & \textbf{1.063} & \textbf{4.295} & 3.630 & 4.472 & 9.264 & {\ul{9.085}} & 2.264 & {\ul{1.715}} & 6.289 \\
 & B & \textbf{1.433} & \textbf{6.137} & {\ul{1.528}} & {\ul{4.397}} & {\ul{3.412}} & \textbf{4.137} & {\ul{7.766}} & 9.190 & \textbf{1.366} & \textbf{1.351} & {\ul{5.496}} \\
\multirow{-3}{*}{\begin{tabular}[c]{@{}c@{}}$r_{err}$\\ (º/100m)\end{tabular}} & C & 2.076 & {\ul{6.231}} & 1.660 & 7.540 & \textbf{1.733} & {\ul{4.157}} & \textbf{6.147} & \textbf{6.299} & {\ul{1.977}} & 1.889 & \textbf{4.595} \vspace{0.6mm} \\ \hline
 & A & 62.718 & 218.055 & \textbf{37.431} & \textbf{8.696} & 3.998 & {\ul{60.149}} & 84.704 & {\ul{35.929}} & 47.987 & {\ul{27.426}} & 26.291 \\
 & B & \textbf{19.424} & {\ul{152.154}} & 80.454 & {\ul{11.412}} & {\ul{3.971}} & \textbf{51.414} & \textbf{69.889} & 36.430 & \textbf{14.708} & \textbf{15.553} & {\ul{23.749}} \\
\multirow{-3}{*}{\begin{tabular}[c]{@{}c@{}}ATE\\ (m)\end{tabular}} & C & {\ul{48.945}} & \textbf{133.490} & {\ul{48.249}} & 15.583 & \textbf{2.469} & 68.946 & {\ul{73.715}} & \textbf{30.908} & {\ul{37.251}} & 27.547 & \textbf{21.791} \vspace{0.6mm} \\ \hline
 & A & 0.107 & \textbf{0.601} & {\ul{0.026}} & \textbf{0.112} & {\ul{0.104}} & {\ul{0.141}} & 0.366 & {\ul{0.164}} & \textbf{0.030} & 0.037 & \textbf{0.146} \\
 & B & \textbf{0.037} & 1.060 & \textbf{0.024} & 0.138 & 0.113 & {\ul{0.141}} & \textbf{0.290} & 0.178 & {\ul{0.031}} & {\ul{0.033}} & {\ul{0.150}} \\
\multirow{-3}{*}{\begin{tabular}[c]{@{}c@{}}RPE\\ (m)\end{tabular}} & C & {\ul{0.038}} & {\ul{0.646}} & 0.040 & {\ul{0.130}} & \textbf{0.066} & \textbf{0.140} & {\ul{0.307}} & \textbf{0.157} & 0.041 & \textbf{0.031} & 0.166 \vspace{0.6mm} \\ \hline
 & A & 0.361 & 0.320 & 0.283 & {\ul{0.288}} & 0.170 & {\ul{0.263}} & 0.252 & {\ul{0.289}} & 0.281 & 0.229 & {\ul{0.324}} \\
 & B & {\ul{0.356}} & {\ul{0.299}} & {\ul{0.277}} & \textbf{0.285} & {\ul{0.165}} & \textbf{0.262} & {\ul{0.241}} & 0.290 & {\ul{0.274}} & {\ul{0.221}} & 0.327 \\
\multirow{-3}{*}{\begin{tabular}[c]{@{}c@{}}RPE\\ (º)\end{tabular}} & C & \textbf{0.339} & \textbf{0.296} & \textbf{0.267} & {\ul{0.288}} & \textbf{0.148} & {\ul{0.263}} & \textbf{0.237} & \textbf{0.284} & \textbf{0.258} & \textbf{0.219} & \textbf{0.321} \vspace{0.6mm} \\ \hline
\end{tabular}
}
\end{table*}

Note that the results of sequences 00, 02, 08, and 09 should not be considered in the analysis once they are used for training the models. However, they are shown in Table~\ref{tab:results} for a complete report, as well as to confirm that the models are not completely overfitted to the training data. 

By ignoring those sequences, it can be seen that the presence of the motion consistency loss in models B and C increased the overall performance if compared to model A (only the MSE loss). Considering the $t_{err}$, model B outperforms model A in all sequences but in sequences 01 and 03. For the ATE, model B showed worse performance than model A only in sequence 03. However, this metric reveals that the consistency loss led to a significant improvement, especially in sequences 01, 05, and 06. Similar conclusions can be made for the rotational metric. For example, the $r_{err}$ shows that model B outperforms model A in all sequences but in sequences 03 and 07. However, the rotational metrics were not significantly affected by the consistency loss, once the metric values are close to each other for all the models. The same happens for the relative pose errors. Furthermore, similar conclusions can be drawn for model C since it is also frequently present in the top 1 and top 2 best values.

Therefore, the consistency loss improved the overall performance when comparing models B and C with model A, especially regarding the translational aspect.

\begin{figure*}[!t]
\centering
\includegraphics[width=0.78\textwidth]{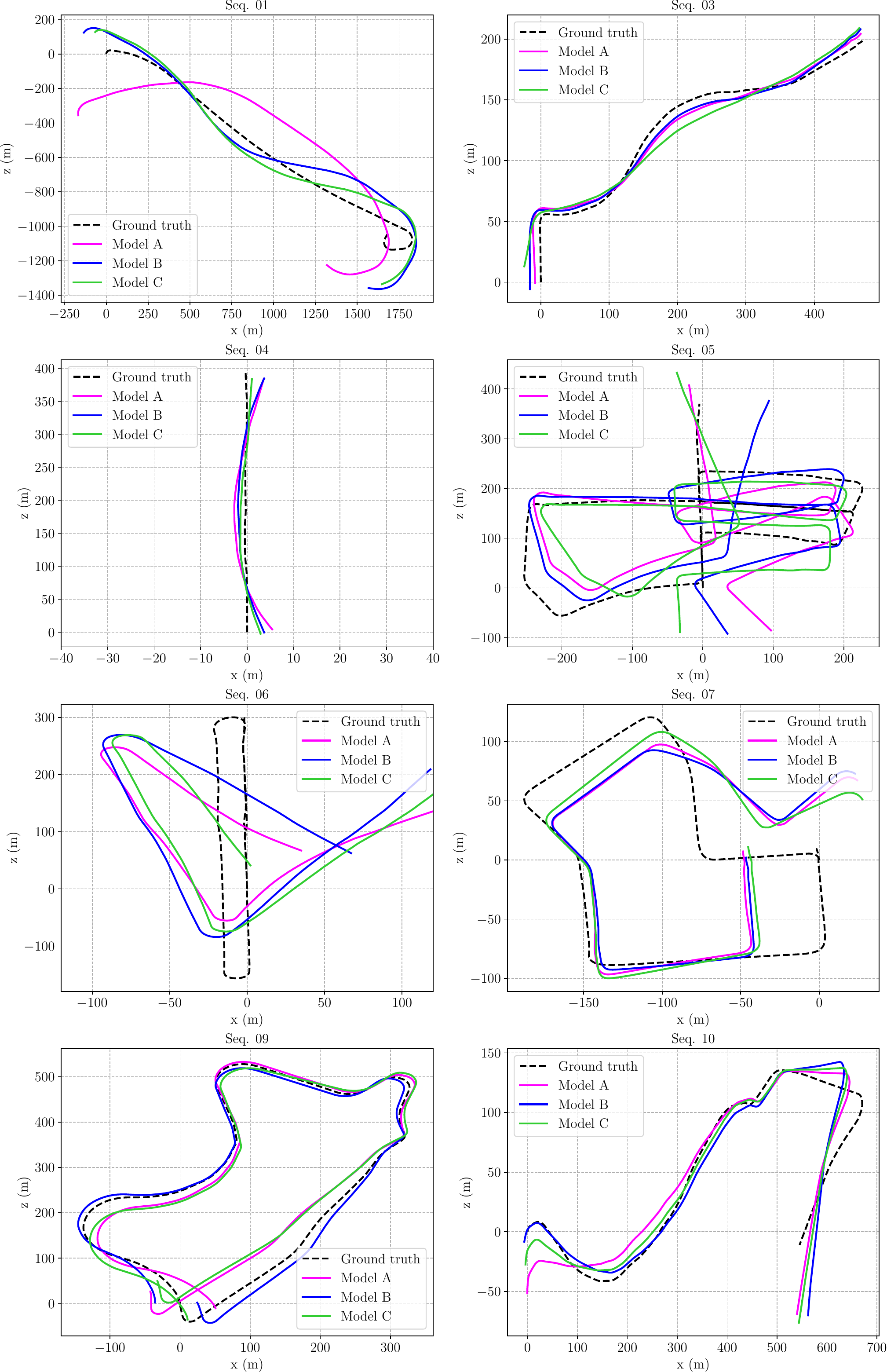}
\caption{Trajectories obtained by model A(\colorindicator{tab:magenta}), model B (\colorindicator{tab:b}), and model C (\colorindicator{tab:limegreen}), compared with the ground truth (\colorindicator{tab:k}). All depicted sequences belong to the test set, but sequence 09. Trajectories are obtained under the 7-DoF alignment.}
\label{fig:traj_results}
\end{figure*}

Following the analysis of the impact of the consistency loss in visual odometry, we displayed in Fig.~\ref{fig:traj_results} the predicted trajectories of models A, B, and C for qualitative analysis. Fig.~\ref{fig:traj_results} reveals that model B is predominantly closer to the ground truth than the other models. Moreover, it also reveals that the scale drift problem still remains pronounced, typical of monocular systems.

\section{Conclusion}
\label{sec:conclusion}

We proposed a consistency loss for visual odometry systems with deep learning. Experimental results have demonstrated that the addition of our motion consistency loss increased the overall performance of the model considering the KITTI dataset. Our results also showed that the consistency loss was most prominent for the translational metrics, not being highly effective in the rotational aspect. Despite an improvement in model performance for visual odometry, the scale drift problem still remains significant for monocular systems, which is expected since the model is not designed to infer depth features to estimate the scale. 

For future research, we expect to minimize the scale drift problem by addressing a depth estimator together \cite{francani2022dense} with our approach. Furthermore, we are looking for new deep learning-based architectures to estimate the camera's pose in an end-to-end manner. 

\bibliographystyle{IEEEtran}
\bibliography{IEEEabrv, refs}

\end{document}